\documentclass{bmvc2k}

\newcommand\etc{etc\@ifnextchar.{}{.\@}}
\usepackage{amssymb}
\usepackage[table]{xcolor}
\usepackage{floatrow}
\newfloatcommand{capbtabbox}{table}[][\FBwidth]

\title{One-Shot Learning for Semantic Segmentation}

\addauthor{Amirreza Shaban}{amirreza@gatech.edu}{1}
\addauthor{Shray Bansal}{sbansal34@gatech.edu}{1}
\addauthor{Zhen Liu}{liuzhen1994@gatech.edu}{1}
\addauthor{Irfan Essa}{irfan@gatech.edu}{1}
\addauthor{Byron Boots}{bboots@cc.gatech.edu}{1}

\addinstitution{
 College of Computing,\\
 Georgia Institute of Technology,\\
 Georgia, USA
}

\runninghead{Shaban, Bansal, Liu, Essa, Boots}{One-Shot Semantic Segmentation}

\def\etal{\emph{et al}\bmvaOneDot}
\def\etc{\emph{etc}\bmvaOneDot}
\newcommand{\plh}{%
  {\ooalign{$\phantom{0}$\cr\hidewidth$\scriptstyle\times$\cr}}%
}

\begin{document}

\maketitle

\begin{abstract}
Low-shot learning methods for image classification support learning from sparse data. We extend these techniques to support dense semantic image segmentation. Specifically, we train a network that, given a small set of annotated images, produces parameters for a Fully Convolutional Network (FCN). We use this FCN to perform dense pixel-level prediction on a test image for the new semantic class. Our architecture shows a 25\% relative meanIoU improvement compared to the best baseline methods for one-shot segmentation on unseen classes in the PASCAL VOC 2012 dataset and is at least $3\plh$  faster. The code is publicly available at: \url{https://github.com/lzzcd001/OSLSM}.
\end{abstract}

\section{Introduction}
Deep Neural Networks are powerful at solving classification problems in computer vision. However, learning classifiers with these models requires a large amount of labeled training data, and recent approaches have struggled to adapt to new classes in a data-efficient manner.  
There is interest in quickly learning new concepts from limited data using one-shot learning methods~\cite{kaiser2017learning, santoro2016meta}. One-shot image classification is the problem of classifying images given only a single training example for each category~\cite{vinyals2016matching,koch2015siamese}.

We propose to undertake \textit{One-Shot Semantic Image Segmentation}. Our goal is to predict a pixel-level segmentation mask for a semantic class (like horse, bus, \etc) given only a single image and its corresponding pixel-level annotation. We refer to the image-label pair for the new class as the support set here, but more generally for $k$-shot learning, support set refers to the $k$ images and labels.

%

\begin{figure}[t]
\begin{center}
\includegraphics[width=0.8\linewidth]{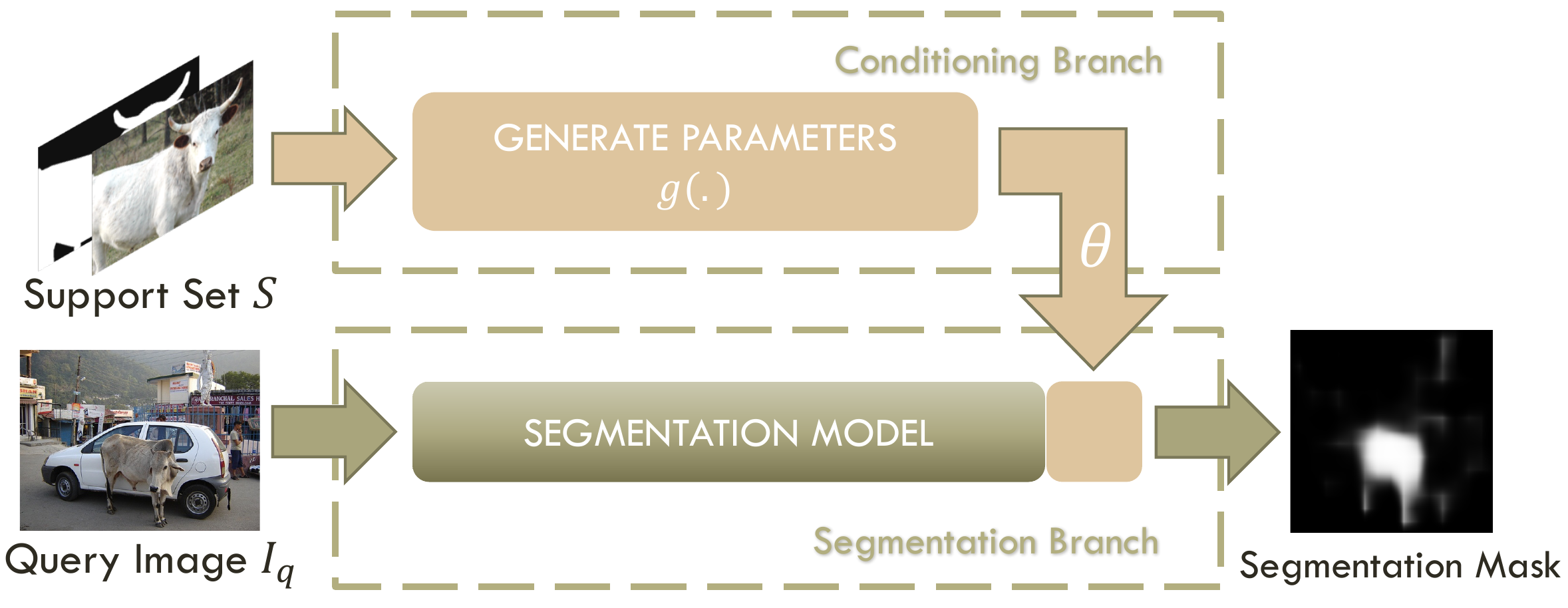}
\end{center}
   \caption{Overview. S is an annotated image from a new semantic class. 
In our approach, we input S to a function $g$ that outputs a set of parameters $\theta$. We use $\theta$ to parameterize part of a learned segmentation model which produces a segmentation mask given $I_q$.}
\label{fig:problem_setup}
\end{figure}

A simple approach to performing one-shot semantic image segmentation is to fine-tune a pre-trained segmentation network on the labeled image~\cite{caelles2016one}. This approach is prone to over-fitting due to the millions of parameters being updated. It also introduces complications in optimization, 
where parameters like step size, momentum, number of iterations, \etc may be difficult to determine. 
Recent one-shot image categorization methods~\cite{vinyals2016matching,koch2015siamese} in contrast, meta-learn a classifier that, when conditioned on a few training examples, can perform well on new classes.
Since Fully Convolutional Neural Networks (FCNs)~\cite{long2015fully} perform segmentation as pixel-wise classification, we could extend these one-shot methods directly to classify at the pixel level. However, thousands of dense features are computed from a single image and one-shot methods do not scale well to this many features. We illustrate this issue by implementing an extension to the Siamese Network from \cite{koch2015siamese} as a baseline in Section \ref{sec:baselines}.    

We take inspiration from few-shot learning and propose a novel two-branched approach to one-shot semantic image segmentation. The first branch takes the labeled image as input and produces a vector of parameters as output. The second branch takes these parameters as well as a new image as input and produces a segmentation mask of the image for the new class as output. This is illustrated in Figure \ref{fig:problem_setup}. 
Unlike the fine tuning approach to one-shot learning, 
which may require many iterations of SGD to learn parameters for the segmentation network, the first branch of our network computes parameters in a single forward pass. This has several advantages: the single forward pass makes our method fast; our approach for one-shot learning is fully differentiable, allowing the branch to be jointly trained with the segmentation branch of our network; 
finally, the number of parameters $\theta$ is independent of the size of the image, so our method does not have problems in scaling.

To measure the performance for one-shot semantic segmentation we define a new benchmark on the PASCAL VOC 2012 dataset~\cite{pascal-voc-2012} (Section \ref{sec:dataset}). The training set contains labeled images from a subset of the PASCAL classes and the testing set has annotations of classes that were not present in training. We show significant improvements over the baselines on this benchmark in terms of the standard meanIoU (mean Intersection over Union) metric as described in Section~\ref{sec:experiments}. 

We extend to $k$-shot learning by applying our one-shot approach for each of the $k$ images independently to produce $k$ segmentation masks. We then aggregate these masks by performing a logical-OR operation at the pixel level. This approach, apart from being easy to implement and fast, requires no retraining to generalize to any number of images in the support set. We show its effectiveness in terms of increasing meanIOU accuracy per added image to the support set in section \ref{sec:experiments}.\ 

PASCAL VOC contains only 20 classes, which is small when compared to standard datasets used for training one-shot classification methods like Omniglot (1623) \cite{lake2015human} and ImageNet (1000) (\cite{imagenet_cvpr09}). 
Simulating the one-shot task during training, even with such a limited number of classes performs well. This is in contrast to the common notion that training models for few-shot learning requires a large number of classes. We hypothesize that part of our algorithm's ability to generalize well to unseen classes comes from the pre-training performed on ImageNet, which contains weak image-level annotations for a large number of classes. 
We perform experiments on the pretraining in section \ref{subsec:pretrain}.

This paper makes the following contributions: (1) we propose a novel technique for one-shot segmentation 
which outperforms baselines while remaining significantly faster; (2) we show that our technique can do this without weak labels for the new classes; (3) we show that meta-learning can be effectively performed even with only a few classes having strong annotations available; and (4) we set up a benchmark for the challenging $k$-shot semantic segmentation task on PASCAL.

\section{Related Work}
\noindent\textbf{Semantic Image Segmentation} is the task of classifying every pixel in an image into a predefined set of categories. Convolutional Neural Network (CNN) based methods have driven recent success in the field. Some of these classify super-pixels~\cite{girshick2014rich, mostajabi2015feedforward, hariharan2014simultaneous}, others classify pixels directly~\cite{long2015fully, chen2016deeplab, hariharan2015hypercolumns, noh2015learning}. 
We base our approach on the Fully Convolutional Network (FCN) for Semantic Segmentation~\cite{long2015fully} which showed the efficiency of pixel-wise classification. However, unlike FCN and the other approaches above, we do not assume a large set of annotated training data for the test classes.


\noindent\textbf{Weak Supervision.} Weak and semi-supervised methods for Semantic Segmentation reduce the requirement on expensive pixel-level annotations, thus attracting recent interest. Weak supervision refers to training from coarse annotations like bounding boxes~\cite{dai2015boxsup} or image labels~\cite{pinheiro2015weakly,papandreou2015weakly,pathak2014fully}. 
A notable example is co-segmentation, where the goal is to find and segment co-occurring objects in images from the same semantic class~\cite{rother2006cosegmentation, faktor2013co}. Many co-segmentation algorithms~\cite{hochbaum2009efficient, chen2014enriching, quan2016object} assume object visual appearances in a batch are similar and either rely on hand-tuned low-level features or high-level CNN features trained for different tasks or objects ~\cite{quan2016object}. In contrast, we meta-learn a network to produce a high-level representation of a new semantic class given a single labeled example. 
Semi-supervised approaches~\cite{papandreou2015weakly,hong2016learning,hong2015decoupled} combine weak labels with a small set of pixel-level annotations. However, they assume a large set of weak labels for each of the desired objects. For instance, Pathak~\etal~\cite{pathak2015constrained} use image-level annotations for all classes and images in the PASCAL 2012 training set~\cite{pascal-voc-2012}, while we exclude all annotations of the testing classes from the PASCAL training set.

\noindent\textbf{Few-Shot Learning} algorithms seek to generalize knowledge acquired through classes seen during training to new classes with only a few training examples~\cite{li2006one, salakhutdinov2012one, vinyals2016matching}. 
Discriminative methods in which the parameters of the base classifier (learned on training classes) are adapted  to the new class~\cite{bart2005cross, HariharanG16, bertinetto2016learning, wang2016learning} are closely related to our work. The main challenge is that the adapted classifier is prone to over-fit to the newly presented training examples. Wang and Herbert~\cite{wang2016learning} address this challenge by learning to predict classifiers which remain close to the base classifier. Bertinetto~\etal~\cite{bertinetto2016learning} trained a two-branch network, in which one branch receives an example and predicts a set of dynamic parameters. The second branch classifies the query image using the dynamic parameters along with a set of learned static parameters. A similar approach was used by Noh~\etal in~\cite{noh2016image} for question answering. We draw several ideas from these papers and adapt them for the task of dense classification to design our model.
Metric learning is another approach to low-shot learning~\cite{vinyals2016matching, koch2015siamese}. It aims to learn an embedding space that pulls objects from the same categories close, while pushing those from different categories apart. Koch~\etal~\cite{koch2015siamese} show that a Siamese architecture trained for a binary verification task can beat several classification baselines in $k$-shot image classification. We adapt their approach for image segmentation as one of our baselines. 

\section{Problem Setup}
\label{section:problem}
Let the support set $S=\{(I_s^i, Y_s^i(l))\}^k_{i=1}$ be a small set of $k$ image-binary mask pairs where $Y_s^i \in L_{test}^{H \times W}$ is the segmentation annotation for image $I_s^i$ and $Y_s^i(l)$ is the mask of the $i^{th}$ image for the semantic class $l \in L_{test}$. The goal is to learn a model $f(I_q, S)$ that, when given a support set $S$ and query image $I_q$, predicts a binary mask $\hat{M_q}$ for the semantic class $l$. An illustration of the problem for $k=1$ is given Figure~\ref{fig:problem_setup}.

During training, the algorithm has access to a large set of image-mask pairs $D = \{(I^{j}, Y^{j})\}^N_{j=1}$ where $Y^{j} \in L_{train}^{H \times W}$ is the semantic segmentation mask for training image $I^{j}$. 
At testing, the query images are only annotated for new semantic classes i.e. $L_{train} 
\cap L_{test} = \varnothing$
. This is the key difference from typical image segmentation where training and testing classes are the same. 
While the problem is similar to $k$-shot learning, which has been extensively studied for image classification \cite{salakhutdinov2012one, vinyals2016matching}, applying it to segmentation requires some modification. 

In this problem, unlike image classification, examples from $L_{{test}}$ might appear in training images. This is handled naturally when an annotator unaware of some object class, labels it as background. 
Annotations of $L_{{test}}$ objects are excluded from the training set, while the images are included as long as there is an object from $L_{{train}}$ present. 
State-of-the-art algorithms for image segmentation~\cite{chen2016semantic, chen2014semantic} use networks pre-trained on large-scale image classification datasets like ~\cite{imagenet_cvpr09}. 
Although these weights give the models a better starting point, they still require many segmented images and thousands of weight updates to learn a good model for pixel classification. This is true even for the classes that directly overlap. We allow similar access to weak annotations for our problem by initializing VGG with weights pre-trained on ImageNet \cite{imagenet_cvpr09}. In section \ref{subsec:pretrain} however, we show that even excluding all the overlapping classes from pre-training does not degrade the performance of our approach.

\section{Proposed Method}
\label{section:proposed}

\begin{figure*}
  \begin{center}
  \includegraphics[width=.8\textwidth]{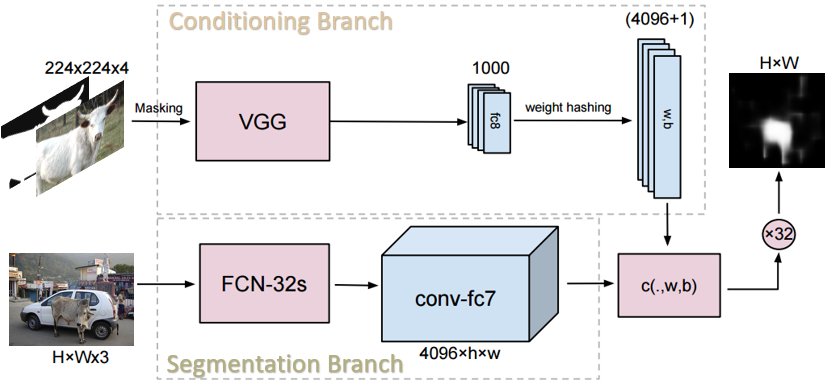}
  \end{center}
\caption{Model Architecture. The conditioning branch receives an image-label pair and produces a set of parameters $\{w,b\}$ for the logistic regression layer $c(\cdot, w, b)$. The segmentation branch is an FCN that receives a query image as input and outputs strided features of conv-fc7. The predicted mask is generated by classifying the pixel-level features through $c(\cdot, w, b)$, which is then upsampled to the original size.}
\label{fig:network}
\end{figure*}

We propose an approach where the first branch receives as input a labeled image from the support set $S$ and the second branch receives the query image $I_q$. In the first branch, we input the image-label pair $S = (I_s, Y_s(l))$ to produce a set of parameters,
\begin{equation}
w, b = g_\eta(S).
\end{equation}
In the other branch, we extract a dense feature volume from $I_q$ using a parametric embedding function $\phi$. Let $F_q = \phi_\zeta(I_q)$ be that feature volume extracted from $I_q$, then $F_q^{mn}$ is the feature vector at the spatial location $(m,n)$. 
Pixel level logistic regression is then performed on the features using the parameters from the first layer to get the final mask,
\begin{equation}
\hat{M}_q^{mn} = 
\sigma({w}^\top F_q^{mn} + b).
\end{equation}
Here, $\sigma(.)$ is the sigmoid function and $\hat{M}_q^{mn}$ is the $(m,n)$ location of the predicted mask for the query. 
This can be understood as a convolutional layer with parameters $\{w,b\}$ followed by a sigmoid activation function, where the parameters are not fixed after training and get computed through the first branch for each image in the support set. The predicted mask is then upsampled back to the original image size using standard bilinear interpolation. The final binary mask is produced by using a threshold of $0.5$ on $\hat{M_q}$. The overall architecture is illustrated in Figure \ref{fig:network}. We explain each part of the architecture in more detail in the following subsections.

\subsection{Producing Parameters from Labeled Image}
We modify the VGG-16 architecture from ~\cite{Simonyan14c} to model the function $g_{\eta}(\cdot)$.

\textbf{Masking. } 
We chose to mask the image with its corresponding label so it contains only the target object instead of modifying the first layer to receive the four channel image-mask pair as input. We do this for the following two empirical reasons. (1) Even in the presence of the mask the network response tends to be biased towards the largest object in the image which may not be the object we would like to segment. (2) Including the background information in the input increased the variance of the output parameters $\{w,b\}$ which prevented the network from converging.

\textbf{Weight Hashing. }Inspired by Noh~\etal~\cite{noh2016image}, we employed the weight hashing layer from \cite{chen2015compressing} to map the $1000$-dimensional vector output from the last layer of VGG to the $4097$ dimensions of $\{w,b\}$. This mapping avoids the overfitting which would occur due to the massive number of extra parameters that a fully connected layer will introduce if used instead. We implemented it efficiently as a fully connected layer with fixed weights. This is explained in more detail in the supplementary material.

\subsection{Dense Feature Extraction}
We model the embedding function $F_q = \phi_\zeta(I_q)$ by the FCN-$32$s fully convolutional architecture~\cite{long2015fully} excluding the final prediction layer. The $4096$ channel feature volume at conv-fc7 is then fed to the logistic pixel classifier described above. In section \ref{sec:experiments} we also evaluate performance of the high resolution dilated-FCN~\cite{yu2015multi} with stride $8$.

\subsection{Training Procedure}
\label{section:training}
We simulate the one shot task during training by sampling a support set $S$, a query image $I_q$ and its corresponding binary mask $M_q$ from the training set $D_{train}$ at each iteration.
First, an image-label pair $(I_q, Y_q)$ is sampled uniformly at random from $D_{train}$, then we sample a class $l \in$ $L_{{train}}$ uniformly from the classes present in the semantic mask and use it to produce the binary mask $Y_q(l)$. $S$ is formed by picking one image-mask pair at random from $D_{{train}} - \{(I_q, Y_q)\}$ with class $l$ present. 
We can then predict the mask $\hat{M}_q$ with a forward pass through our network. We maximize the log likelihood of the ground-truth mask
\begin{equation}
\label{eq:loss_function}
\mathcal{L}(\eta, \zeta) = \mathop{\mathbb{E}}_{S, I_q, M_q \sim D_{train}} \Big[ \sum_{m,n} \log p_{\eta, \zeta}(M_q^{mn} | I_q, S) \Big].
\end{equation}
Here $\eta$ and $\zeta$ are the network parameters, $p_{\eta, \zeta}$ is the probability of the mask given the neural network output, and $S$, $I_q$, and $M_q$ are sampled by the sampling strategy described above. We use Stochastic Gradient Descent with a fixed learning rate of $10^{-10}$, momentum $0.99$ and batch size of $1$. The VGG network overfits faster than the fully-convolutional branch; therefore, we set the learning rate multiplier to $0.1$ for learning the parameter $\eta$. We stop training after $60$k  iterations.

\subsection{Extension to $k$-shot}
In the case of $k$-shot segmentation the support set contains $k$ labeled images, $S = \{I_s^i, Y_s^i(l)\}_{i=1}^k$. We use these images to produce $k$ sets of the parameters $\{w^i, b^i\}_{i=1}^k$. Each of them can be understood to be an independent classifier of an ensemble. These classifiers we noticed have high precision but low recall. We believe this is because each is produced by one example from the support set and a single image can only contain a small subset of the possible appearances of the object. So, we combine the decision of these classifiers by including a pixel in the final mask if it was considered an object by any of these classifiers. This is implemented as a logical OR operation between the $k$ binary masks. This approach has the benefit that it does not require any retraining and can be generalized to any $k$. It is also much faster than the baselines as shown in section \ref{sec:experiments}.

\section{Dataset and Metric}
\label{sec:dataset}
\textbf{Dataset:} We create a new dataset, \textbf{PASCAL-}$\mathbf{5^i}$, for the problem of $k$-shot Image Segmentation using images and annotations from PASCALVOC 2012 \cite{pascal-voc-2012} and extended annotations from SDS\footnote{For creating the training set, we only include images that do not overlap with the PASCALVOC 2012 validation set.} \cite{hariharan2014simultaneous}. From $L$, the set of twenty semantic classes in PASCALVOC, we sample five and consider them as the test label-set $L_{{test}} = \{4i+1, \dots, 4i+5\}$, with $i$ being the fold number, and the remaining fifteen forming the training label-set $L_{{train}}$. Test and training class names are shown in Table~\ref{table:pascal_names}. We form the training set $D_{{train}}$ by including all image-mask pairs from PASCALVOC and SDS training sets that contain at least one pixel in the semantic mask from the label-set $L_{{train}}$. The masks in $D_{{train}}$ are modified so that any pixel with a semantic class $\neq L_{{train}}$ is set as the background class $l_\varnothing$. We follow a similar procedure to form the test set $D_{{test}}$, but here the image-label pairs are taken from PASCALVOC validation set and the corresponding label-set is $L_{test}$. Thus, apart from a few exclusions, the set of images is similar to those used in Image Segmentation papers, like FCN~\cite{long2015fully}. However, the annotations are different. Given the test set $D_{{test}}$, we use the same procedure that is described in Section \ref{section:training} to sample each test example $\{S, (I_q, Y_q(l))\}$. We sample $N = 1000$ examples and use it as the benchmark for testing each of the models described in the next section.

\noindent\textbf{Metric:} Given a set of predicted binary segmentation masks $\{\hat{M_q}\}_{i=1}^N$ and the ground truth annotated mask $\{M_q\}_{i=1}^N$ for a semantic class $l$ we define the per-class Intersection over Union ($IoU_l$) as $\frac{\mathit{tp}_l}{\mathit{tp}_l + \mathit{fp}_l + \mathit{fn}_l}$.
Here, $\mathit{tp}_l$ is the number of true positives, $\mathit{fp}_l$ is the number of false positives and $\mathit{fn}_l$ is the number of false negatives over the set of masks. The \textbf{meanIoU} is just its average over the set of classes, i.e. $(1 / n_l)\sum_l IoU_l$. This is the standard metric of meanIU defined in Image Segmentation literature adapted for our binary classification problem.
\begin{table*}[t]
\centering
\resizebox{\textwidth}{!}{\begin{tabular}{|c|c|c|c|}
\hline
$i = 0$ & $i = 1$ & $i = 2$ & $i = 3$ \\
\hline\hline
\cellcolor{red!25}aeroplane, bicycle, bird, boat, bottle & \cellcolor{blue!25}bus, car, cat, chair, cow & \cellcolor{green!25}diningtable, dog, horse, motorbike, person & \cellcolor{yellow!25}potted plant, sheep, sofa, train, tv/monitor \\
\hline\end{tabular}}
\label{table:pascal_names}
\centering
\resizebox{.7\textwidth}{!}{\begin{tabular}{l|cccc||c}\\
{\bf Methods ($1$-shot)} & \cellcolor{red!25} PASCAL-$5^0$ & \cellcolor{blue!25}PASCAL-$5^1$ & \cellcolor{green!25}PASCAL-$5^2$ & \cellcolor{yellow!25}PASCAL-$5^3$ & Mean\\ 
\hline\hline
1-NN & $25.3$ & $44.9$ & $\mathbf{41.7}$ & $18.4$ & $32.6$ \\ 
LogReg & $26.9$ & $42.9$ & $37.1$ & $18.4$ & $31.4$ \\ 
Finetuning & $24.9$ & $38.8$ & $36.5$ & $30.1$ & $32.6$\\ 
Siamese & $28.1$ & $39.9$ & $31.8$ & $25.8$ & $31.4$ \\ 
Ours & $\mathbf{33.6}$ & $\mathbf{55.3}$ & $40.9$ & $\mathbf{33.5}$ & $\mathbf{40.8}$ \\ 
\end{tabular}}
\label{table:full_results_k1}
\centering
\resizebox{.7\textwidth}{!}{\begin{tabular}{l|cccc||c}\\
{\bf Methods ($5$-shot)} & \cellcolor{red!25}PASCAL-$5^0$ & \cellcolor{blue!25}PASCAL-$5^1$ & \cellcolor{green!25}PASCAL-$5^2$ & \cellcolor{yellow!25}PASCAL-$5^3$ & Mean\\ 
\hline\hline
Co-segmentation & $25.1$ & $28.9$ & $27.7$ & $26.3$ & $27.1$\\ \hline
1-NN & $34.5$ & $53.0$ & $\mathbf{46.9}$ & $25.6$ & $40.0$\\ 
LogReg & $\mathbf{35.9}$ & $51.6$ & $44.5$ & $25.6$ & $39.3$\\ 
Ours & $\mathbf{35.9}$ & $\mathbf{58.1}$ & $42.7$ & $\mathbf{39.1}$ & $\mathbf{43.9}$\\ 
\end{tabular}}
\caption{Mean IoU results on PASCAL-$5^i$. \textbf{Top:} test classes for each fold of PASCAL-$5^i$. The \textbf{middle} and \textbf{bottom} tables contain the semantic segmentation meanIoU on all folds for the $1$-shot and $5$-shot tasks respectively.}
\label{table:full_results_k5}
\end{table*}
\section{Baselines}
\label{sec:baselines}
We evaluate the performance of our method with different baselines. Since one-shot image segmentation is a new problem, we adapt previous work for dense pixel prediction to serve as baselines to compare against.
\begin{itemize}
\item \textbf{Base Classifiers:} CNNs learn deep representations of images, so these models are an intuitive starting point for classification. Specifically, we first fine-tune FCN-$32$s pretrained on ILSVRC2014 data to perform $16$-way ($15$ training foreground classes + $1$ background class) pixel-wise predictions on the \textbf{PASCAL-}$\mathbf{5^i}$ dataset. During testing, we extract dense pixel-level features from both images in the support set and the query image. We then train classifiers to map dense fc-$7$ features from the support set to their corresponding labels and use it to generate the predicted mask $\hat{M}_{q}$. We experimented with various classifiers including $1$-NN and logistic regression\footnote{We also trained linear SVM, but could not get a comparable results to logistic regression.}
\item \textbf{Fine-tuning:} As suggested by~\cite{caelles2016one}, for each test iteration we fine-tune the trained segmentation network on examples in the support set and test on the query image. We only fine-tune the fully connected layers (fc6, fc7, fc8) to avoid overfitting and reducing the inference time per query. We also found that the fine-tuned network converges faster if we normalize the fc-$7$ features by a batch normalization layer.
\item \textbf{Co-segmentation by Composition:} To compare with the these  techniques, we include the results of the publicly available implementation\footnote{\url{http://www.wisdom.weizmann.ac.il/~vision/CoSegmentationByComposition.html}} of ~\cite{faktor2013co} on PASCAL-5$^i$. 

\item \textbf{Siamese Network for One-shot Dense Matching:} Siamese Networks trained for image verification, i.e. predicting whether two inputs belong to the same class, have shown good performance on one-shot image classification \cite{koch2015siamese}. We adapt them by using two FCNs to extract dense features and then train it for pixel verification. A similarity metric from each pixel in the query image to every pixel in the support set is also learned and pixels are then labeled according to their nearest neighbors in the support set. Implementation details are provided in the supplementary document.
\end{itemize}

\section{Experiments\label{sec:experiments}}
We conduct several experiments to evaluate the performance our approach on the task of $k$-shot Image segmentation by comparing it to other methods. 
Table~\ref{table:full_results_k5} reports the performance of our method in $1$-shot and $5$-shot settings and compares them with the baseline methods. To fit a $5$-shot Siamese network into memory we sampled from features in the support set with a rate of $0.3$. However, sub-sampling considerably degraded the performance of the method and $5$-shot results were worse than the $1$-shot version so we exclude those results. 

\begin{figure}[t]
\begin{center}
\includegraphics[width=.8\linewidth]{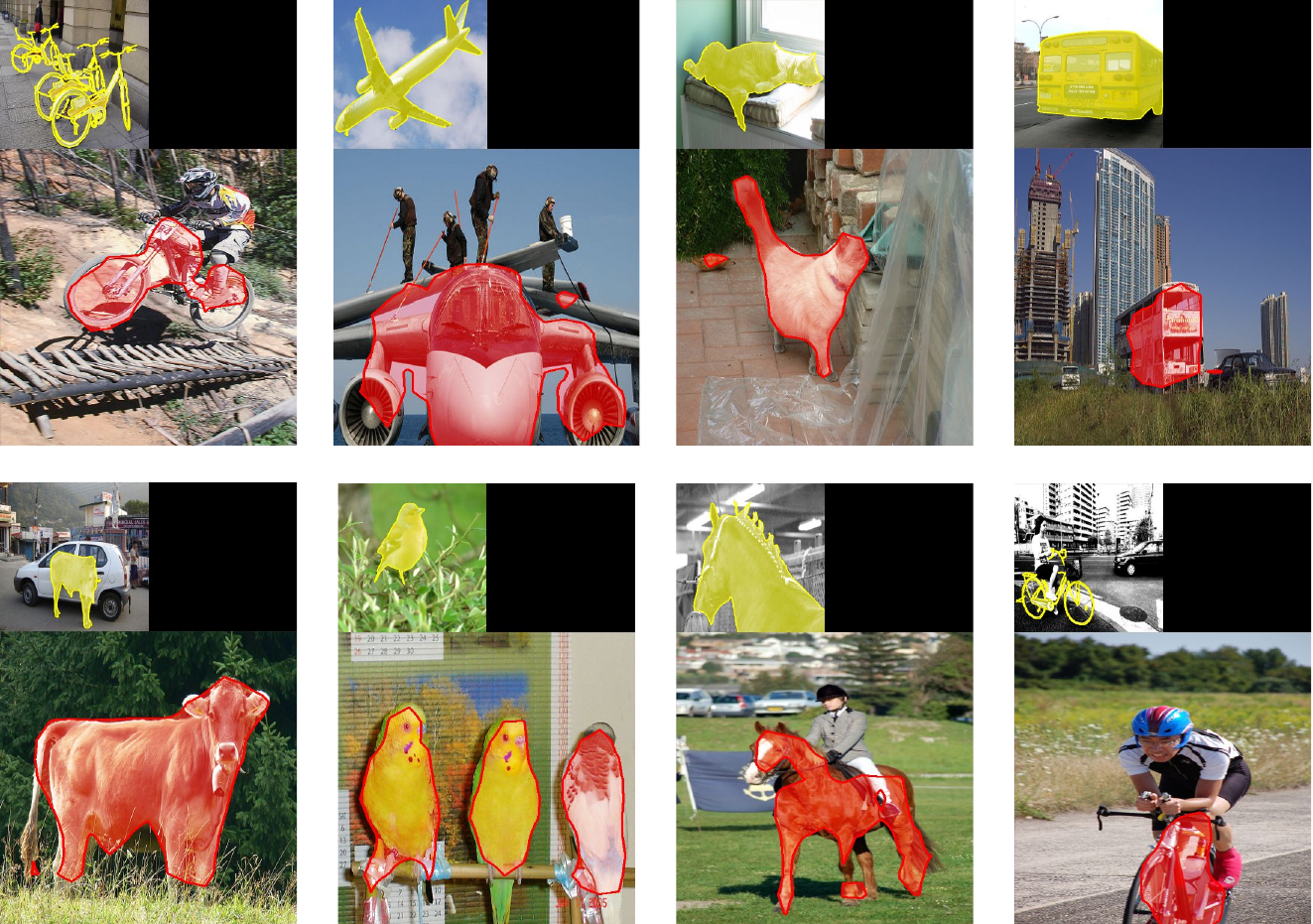}
\end{center}
\caption{Some qualitative results of our method for $1$-shot. Inside each tile, we have the support set at the top and the query image at the bottom. The support is overlaid with the ground truth in yellow and the query is overlaid with our predicted mask in red.}
\label{fig:qual-results}
\end{figure}

Our method shows better generalization performance to new classes. The difference is very noticeable in $1$-shot learning as other methods overfit to only the image in the support set. Specifically, our method outperforms 1-NN and fine-tuning in one-shot image segmentation by 25\% relative meanIoU. We also provide some qualitative result from our method in Figure~\ref{fig:qual-results}. Surprisingly, the results for 1-NN are almost as good as the fine-tuning baseline, which overfits quickly to the data in the support set.

In Table~\ref{table:full_results_k5}, we also compare Co-segmentation by Composition \cite{rother2006cosegmentation} for $5$-shot segmentation to our approach. As expected, using the strong pixel-level annotations enables our method to outperform the unsupervised co-segmentation approach, by $16\%$. In fact, we can  outperform co-segmentation results that require $5$ weakly annotated images with just a single strongly annotated image.

\noindent\textbf{Dilated-FCN:} In addition to the low-resolution version of our method, we also trained the dilated-FCN with higher resolution on $\textbf{PASCAL-}$ $\mathbf{5^0}$ and  
achieved $37.0\%$ and $37.43\%$ meanIoU for $1$-shot and $5$-shot respectively. We notice a $3.4\%$ improvement over low-resolution for one-shot, however, the gap between $1$-shot and $5$-shot is small at this resolution. We believe this is due to our training being specific to the $1$-shot problem. We do not use dilated-FCN architecture for other methods due to the impracticality caused by their high computational cost or memory footprint. 

\begin{figure}
\begin{floatrow}
\ffigbox{%
  \includegraphics[width=1.0\linewidth]{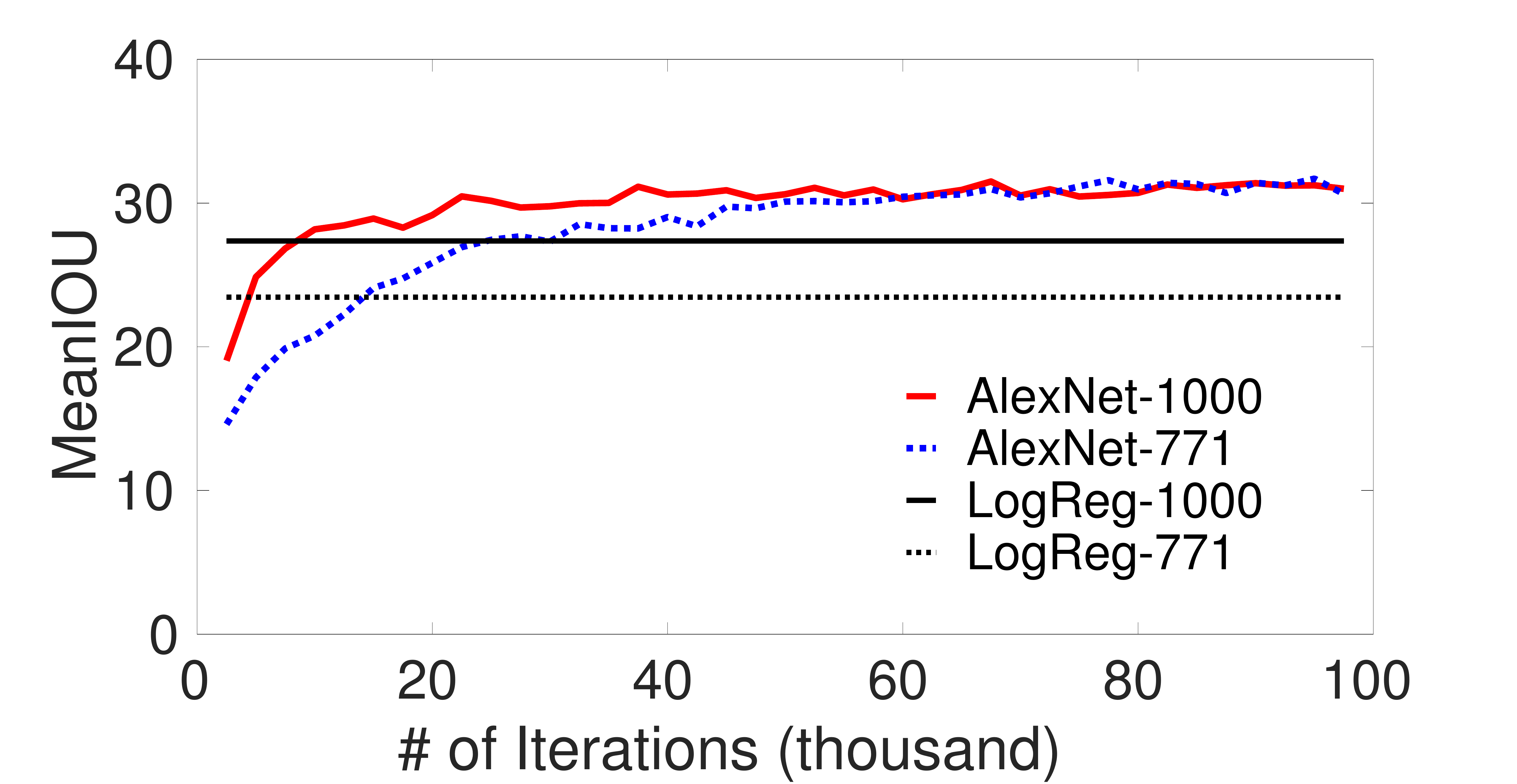}
}{%
  \caption{Pretraining Effect on AlexNet.}
  \label{fig:alexnet-meaniou}
}
\capbtabbox{%
  \begin{tabular}{l|cc}
{\bf Methods} & $1$-shot & $5$-shot \\
\hline\hline
$1$-NN & $1.10$ & $4.55$ \\
Logistic Reg & $0.66$ & $3.50$ \\
Finetune & $5.56$ & - \\
Siamese & $5.65$ & - \\
Ours-32s & $0.19$ & $0.21$
\end{tabular}
}{%
\caption{Inference Time (in s).}
\label{tab:timing}

}
\end{floatrow}
\end{figure}

\noindent\textbf{Running Time:} In Table~\ref{tab:timing} we include the running time of each algorithm. All the experiments were executed on a machine with a $4$GHz Intel Core-i7 
CPU, 32GB RAM, and a Titan X GPU. In one-shot setting our method is $\sim\!3\!\times$ faster the than second fastest method logistic regression. For $5$-shot our method is $\sim\!10\!\times$ faster than logistic regression.

\subsection{Pretraining Effect}
\label{subsec:pretrain}
The models compared above have two sources of information, the image-level labels for the classes in ImageNet \cite{imagenet_cvpr09} through the pretraining and the pixel-level annotation of classes in $L_{{train}}$. Although the test classes $L_{test}$ do not overlap with $L_{{train}}$, they have partial overlap with some ImageNet classes. 
To understand this effect, we use a dataset which excludes all the classes in ImageNet with any overlap with PASCAL categories called PASCAL-removed-ImageNet as in \cite{huh2016makes}. This dataset contains only $771$ classes as compared to $1000$ originally since each class in PASCAL usually overlaps with multiple ImageNet classes. 
We use AlexNet~\cite{NIPS2012_4824} trained on ImageNet and PASCAL-removed-ImageNet (from Huh~\etal~\cite{huh2016makes}) with the suffices $1000$ and $771$ respectively. We replaced the VGG and FCN from both branches of our approach with AlexNet to give us AlexNet-$1000$ and AlexNet-$771$. We also have a baseline in the form of Logistic Regression performed on convolutional AlexNet features finetuned on PASCAL, similar to the Base Classifiers described in section \ref{sec:baselines}. We refer to these as LogReg-$1000$ and LogReg-$771$. Figure~\ref{fig:alexnet-meaniou} contains the results for these models on the first fold, i.e. PASCAL-$5^0$. Note that the results for the two baselines are constant because we evaluate the networks only once they converge.

In Figure~\ref{fig:alexnet-meaniou} we observe that AlexNet-$1000$ is better initially and shows faster convergence. However, after convergence AlexNet-$771$ performs on par with AlexNet-$1000$. The initial gap could be understood by the fact that even the $L_{train}$ classes were not presented during the pre-training. AlexNet being a simpler model performs worse than VGG, meanIOU was $33.6\%$ in Table \ref{table:full_results_k1}. However, AlexNet-$771$ outperforms even our best VGG baseline, which was Siamese at $28.1\%$ for PASCAL-$5^0$. 
This result shows that we can generalize to new categories without any weak supervision for them. In contrast, LogReg-1000 outperforming LogReg-771 shows its incapacity to learn a good representation without seeing weak labels for test categories. This highlights the importance of meta-learning for this task.

\section{Conclusion}
Deep learning approaches have achieved top performance in many computer vision problems. However, learning a new concept given few examples is still a very challenging task. In this paper we developed a new architecture to address this problem for image segmentation. Our architecture learns to learn an ensemble classier and use it to classify pixels in the query image. Through comprehensive experiments we show the clear superiority of our algorithm.  The proposed method is considerably faster than the other baselines and has a smaller memory footprint.

\bibliography{egbib}

\pagebreak
\onecolumn
\appendix
\addcontentsline{toc}{section}{Appendices}
\vspace*{5px}
{\bf\LARGE\color{bmv@sectioncolor} Supplementary Materials}
\vspace*{40px}\\
\noindent\makebox[\linewidth]{\rule{\linewidth}{0.4pt}}
\setcounter{equation}{0}
\setcounter{figure}{0}
\setcounter{table}{0}
\setcounter{page}{1}
\makeatletter
\renewcommand{\theequation}{S\arabic{equation}}
\renewcommand{\thefigure}{S\arabic{figure}}
\renewcommand{\bibnumfmt}[1]{[S#1]}

\section{Weight Hashing}
In our Proposed Work, section 5 of the paper, we employed the weight hashing technique from \cite{chen2015compressing} to map the $1000$-dimensional vector output from the last layer of VGG to the $4097$ dimensions of $\{w,b\}$. This mapping (1) reduces the variance of $\{w,b\}$ as was also noted by Huh \etal in \cite{noh2016image}, and (2) reduces the overfitting which would occur due to the massive number of extra parameters that a fully connected layer will introduce if used instead.

\begin{figure}[h]
\begin{center}
\includegraphics[width=0.35\linewidth]{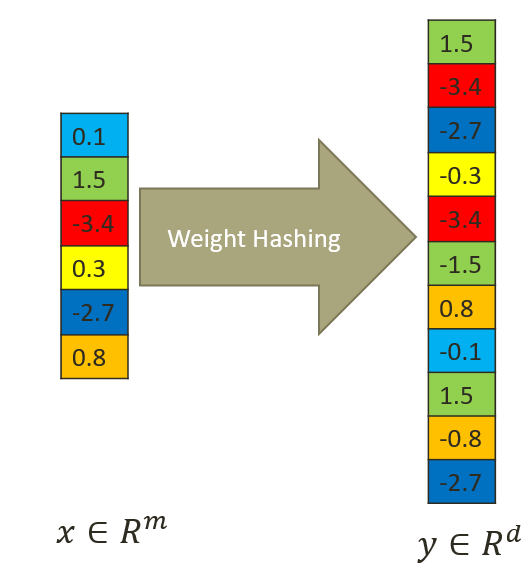}
\end{center}
   \caption{Illustration of weight hashing. In the figure, $x$ is mapped to $y$ by replicating coefficients of $x$ in multiple random locations of $y$ and randomly flipping the sign. The colors help indicate where the entries are copied from.}
\label{fig:weight_hash}
\end{figure} 

Weight hashing is explained as decompression in \cite{chen2015compressing} and is performed as follows. Let $x \in \mathbb{R}^m$ and $\theta \in \mathbb{R}^d$, typically $d > m$, be the inputs and outputs of the layer respectively. Weight hashing works by replicating each coefficient of $x$ in multiple locations of $\theta$ and randomly flipping the sign to reduce the covariance of copied coefficients. Illustrated in Figure \ref{fig:weight_hash}. Specifically, the $i^{th}$ coefficient of $\theta$ is
\begin{equation}
\label{eq:weight_hashing}
\theta(i) = x(p) \zeta(i),
\end{equation}
\begin{equation}
p = \kappa(i),
\end{equation}

where both $\kappa(i) \rightarrow \{1, \dots, m\}$ and $\zeta(i) \rightarrow \{-1, +1\}$ are hashing functions determined randomly. While \cite{chen2015compressing} perform the hashing implicitly to keep the memory footprint small, we implement it as a fully connected layer since we can keep both the hashing values and $\theta$ in memory. The weights are set as 
\begin{equation}
W(i, j) = \zeta(i) \delta_j(\kappa(i)),
\end{equation}
where $\delta_j(\cdot)$ is discreet Dirac delta  function. The weights are set  according to the random hashing functions before training and are kept fixed. This is both easier to implement and more computationally efficient than the original formulation and that used by \cite{noh2016image}. The output of the inner product layer $Wx$ is equal to $\theta$ from Equation~\ref{eq:weight_hashing}. 

\section{Siamese Network for Dense Matching}
\begin{figure}[h]
\begin{center}
\includegraphics[width=0.9\linewidth]{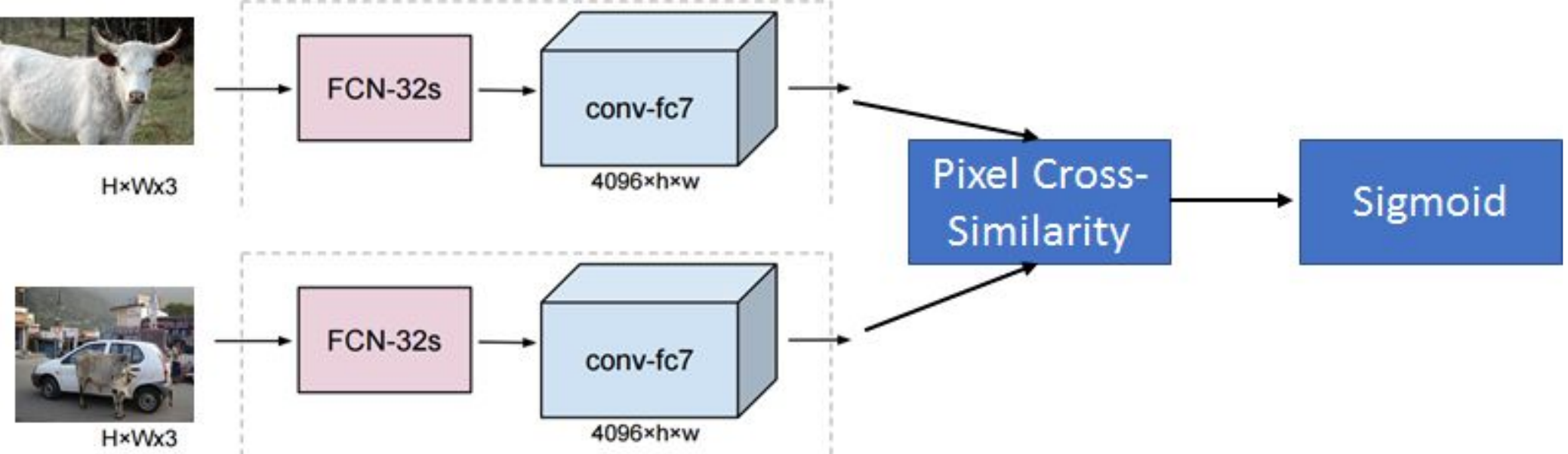}
\end{center}
\caption{Siamese network architeture for dense matching.}
\label{fig:fcn_siamese}
\end{figure}

In the paper, we used the adapted version of Siamese Neural Network for One-shot Image Recognition by Koch \etal \cite{koch2015siamese} for one-shot image segmentation. Here we explain the implementation details. The method from \cite{koch2015siamese} receives as input two images that are each passed through identical convolutional networks and produce a vector as the output for each of them. These vectors are then compared using a learned $L1$ similarity metric and the image is classified according to the label of its nearest neighbor in this metric space. In our case, we use an FCN that outputs a feature volume each for both query and support images. Then the feature for every pixel in the query image is compared to every pixel in the support using a learned $L1$ similarity metric. We implemented this cross similarity measurement between pixels as a python layer for Caffe. The binary label here is assigned to each pixel according to the label of the nearest pixel label in the support set. The whole structure is illustrated in Figure~\ref{fig:fcn_siamese}.

During training, we use FCNs initialized on ImageNet \cite{imagenet_cvpr09} and at each iteration we sample a pair of images from the PASCAL-$5^i$ training set. One of them is treated as the query image and the other becomes the support image. The gradient is computed according to the cross-entropy loss between the sigmoid of the similarity metric and the true binary label. Every batch contains a subset ($50\%$) of the pixels of a query and a support image. Both the similarity metric and the FCN feature extraction are jointly as different parts of the same neural network.   

\section{Qualitative Results}
\begin{figure}[h]
\begin{center}
\includegraphics[width=.8\linewidth]{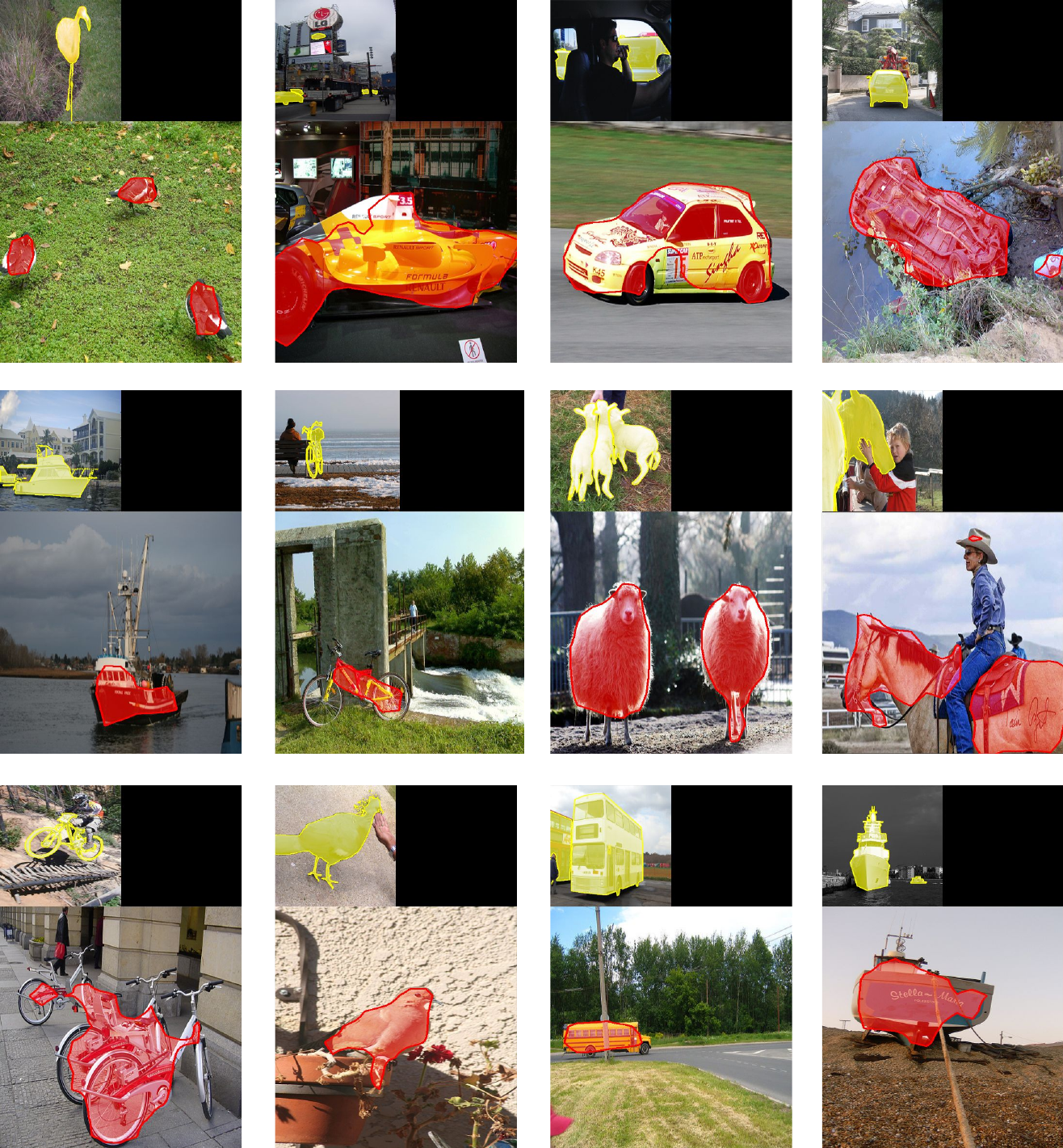}
\end{center}
\caption{Qualitative results for $1$-shot. Inside each tile, we have the support set at the top and the query image at the bottom. The support is overlaid with the ground truth in yellow and the query is overlaid with our predicted mask in red.}
\label{fig:qualitative_supp}
\end{figure}

We include some more qualitative results of our approach for One Shot Semantic Segmentation in Figure \ref{fig:qualitative_supp}. We see that our method is capable of segmenting a variety of classes well and can distinguish an object from others in the scene given only a single support image.  

\begin{figure}[h]
\begin{center}
\includegraphics[width=\linewidth]{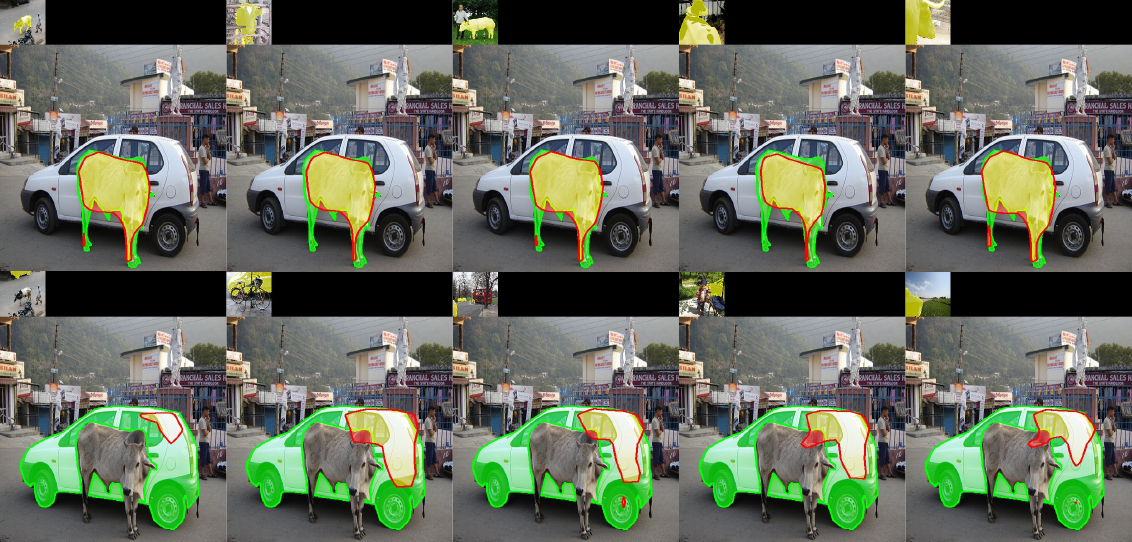}
\end{center}
\caption{Illustration of conditioning effect. Given a fix query image, predicted mask changes by changing the support set. Ground-truth mask is shown in green. First row: support image-mask pairs are sampled from cow class. Second row: support image-mask pairs are sampled from car class. First column: only changing the support mask will will change the prediction.}
\label{fig:conditioning}
\end{figure}

\begin{figure}[h]
\begin{center}
\includegraphics[width=\linewidth]{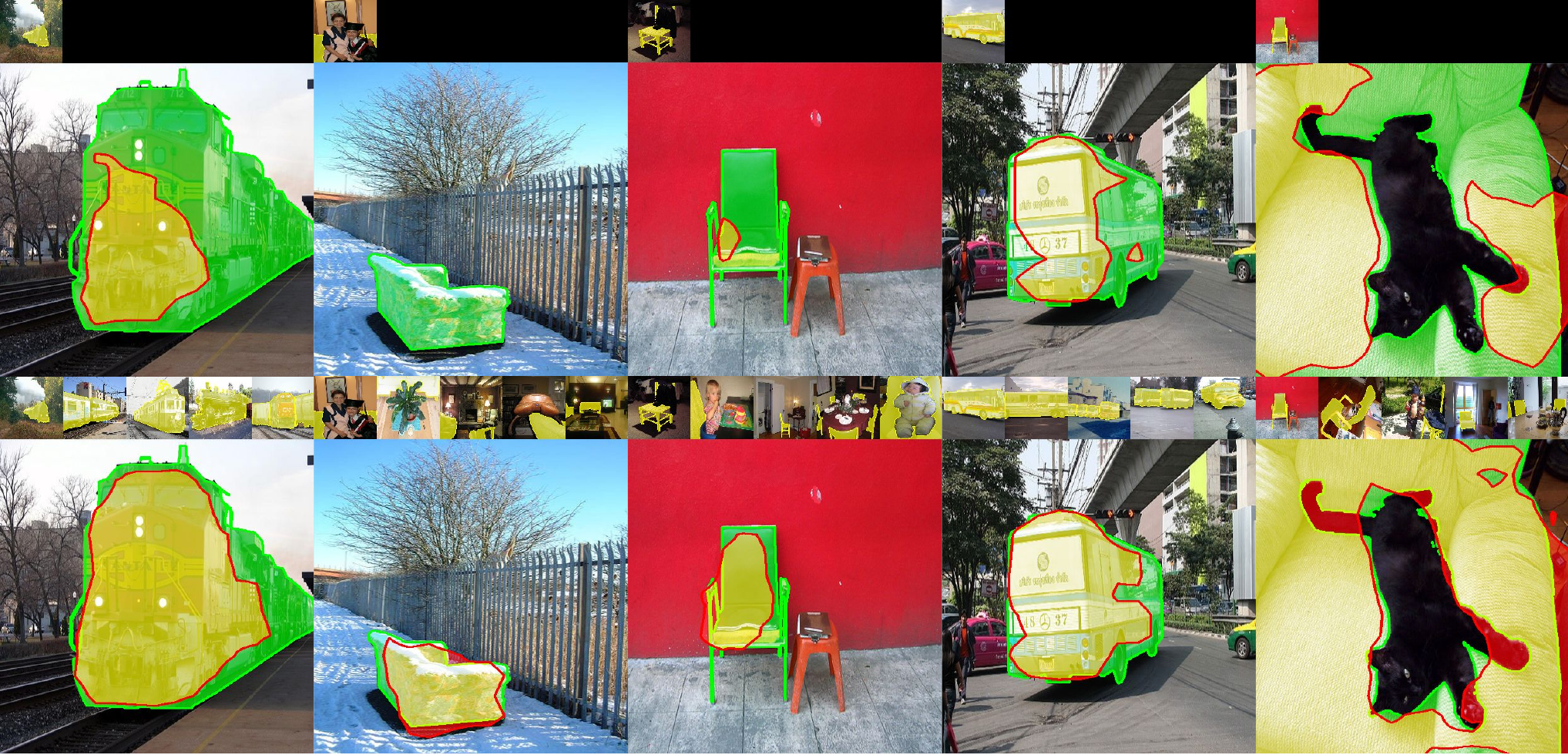}
\end{center}
\caption{Effect of increasing the size of the support set. Results of $1$-shot and 5-shot learning on the same query image are in the first and second rows respectively. Ground truth masks are shown in green and our prediction is in red. The overlap between ground truth and prediction appears yellow.}
\label{fig:1vs5}
\end{figure}

We illustrate the effect of conditioning by segmenting the same query image with different support sets in Figure ~\ref{fig:conditioning}. We picked an unseen query image with two unseen classes, car and cow, and sample support image-mask pairs for each class. Figure~\ref{fig:1vs5} shows how increasing size of the support set helps improve the predicted mask. Note that in both Figures~\ref{fig:1vs5} and~\ref{fig:conditioning} yellow indicates the overlap between ground truth and the prediction. 

\end{document}